\documentclass{article}

\usepackage[final,nonatbib]{nips_2016}

\usepackage[utf8]{inputenc} 
\usepackage[T1]{fontenc}    
\usepackage{hyperref}       
\usepackage{url}            
\usepackage{booktabs}       
\usepackage{amsfonts}       
\usepackage{nicefrac}       
\usepackage{microtype}      
\usepackage{graphicx}
\usepackage{tabularx}
\usepackage{multicol}

\usepackage{array}
\newcolumntype{L}[1]{>{\raggedright\let\newline\\\arraybackslash\hspace{0pt}}m{#1}}
\newcolumntype{C}[1]{>{\centering\let\newline\\\arraybackslash\hspace{0pt}}m{#1}}
\newcolumntype{R}[1]{>{\raggedleft\let\newline\\\arraybackslash\hspace{0pt}}m{#1}}

\usepackage{algorithm}
\usepackage{algpseudocode}
\usepackage{capt-of}
\usepackage{amsmath,amssymb,amsthm,mathrsfs,amsfonts,dsfont}

\newcommand{\E}{\mathds{E}}
\newcommand{\Lo}{\mathcal{L}}
\newcommand{\lo}{\ell}

\title{Gossip training for deep learning}

\author{
  Michael Blot$^{(1)}$ \thanks{Thank to \emph{DGA} for funding.} 
  \And
  David Picard$^{(2)}$ 
  \And
  Matthieu Cord$^{(1)}$
  \And
  Nicolas Thome$^{(1)}$\\
  \AND
  \\
  (1) Sorbonne Universités, UPMC Univ Paris 06, LIP6, 4 place Jussieu 75005 Paris, France \\
  (2) ETIS, ENSEA/Université Cergy-Pontoise/CNRS UMR 8051, 95000 Cergy-Pontoise, France
}

\begin{document}

\maketitle

\begin{abstract}
We address the issue of speeding up the training of convolutional networks. Here we study a distributed method adapted to stochastic gradient descent (SGD). The parallel optimization setup uses several threads, each applying individual gradient descents on a local variable. We propose a new way to share information between different threads inspired by gossip algorithms and showing good consensus convergence properties. Our method called GoSGD has the advantage to be fully asynchronous and decentralized. We compared our method to the recent EASGD in \cite{elastic} on CIFAR-10 show encouraging results.
\end{abstract}

\section{Introduction}
        With deep convolutional neural networks (CNN) introduced by \cite{nakamura} and \cite{lecun}, computer vision tasks and more specifically image classification have made huge improvements last few years following \cite{alexnet}. CNN performances benefit a lot from big databases of annotated images like \cite{imagenet} or \cite{coco}. They are trained by optimizing a loss function with gradient descents computed on random mini-batches. This method called stochastic gradient descent [SGD] has proved to be very efficient to train neural networks in general. 
        
        However current CNN structures are very deep like the 200 layers network ResNet of \cite{resnet} and contains a lot of parameters (around 60M for alexnet\cite{alexnet}) making the training on big datasets very slow. Computation on GPU accelerates the training but it is still difficult to test many architectures. 
        
        Nevertheless the mini-batch optimization seems suitable for distributing the training. Many methods have been proposed like \cite{elastic} or \cite{downpour}. They process SGD in parallel on different threads to optimize a local neural network. The different threads are called workers. Additionally the workers periodically exchange information with a central network. The role of this central variable is essential to share spread information as well as ensuring that all worker networks converge toward a same local minimum. Indeed because of the symmetry property of neural networks well studied in \cite{symmetry} the different workers could give very different optimizations. Having a consensus is important to fully benefit from the parallelism and the information sharing. Unfortunately the proposed methods are not decentralized resulting in loss of time for synchronizing the updates of the central network. It could result of a suboptimal use of distributed computation. 
        
        A well known example of decentralized distributed algorithm is gossip averaging. As studied in \cite{gossipAveraging} this method is very fast to make different agents converge toward a consensus by exchanging information in a peer to peer way. Gossip averaging has already been adapted to other machine learning algorithms such as kernel methods \cite{colin} or PCA \cite{felus}. This family of algorithm presents many advantages like being fully asynchronous and totally decentralized as they do not require a central variable. We propose here to associate this method with SGD in order to apply it to deep learning and more specifically CNN. We call the resulting optimization method GoSGD for Gossip Stochastic Gradient Descent. 
        
        The first section introduces the GoSGD algorithm. Then some experiments illustrate the good convergence properties of decentralized GoSGD.
    
\section{Gossip Stochastic Gradient Descent}
    The objective is to minimize $\Lo(x) = \E_{Y \sim \mathcal{I}}[\lo(x, Y)]$ where $Y$ is a couple variable (image, label) following the natural image distribution, $x$ is the CNN parameters and $\lo$ the loss function. As used in \cite{elastic} and discussed in \cite{distributed} the problem can be derived in a distributed fashion as minimizing:
    \begin{align}
        \label{lossfunction1}
        \sum_{i=1}^M &\Lo(x_i) + \frac{\rho}{2} ||x_i - \overline{x}||_2^2
    \end{align}
    with the $x_i$ being worker's local variables and $\overline{x} = \frac{1}{M}\sum_{i=1}^M x_i$ the global consensus.\\
    We can rewrite this loss in order to exhibit gossip exchanges:    
    \begin{align}
        \label{lossfunction2}
        \sum_{i=1}^M &\Lo(x_i) + \frac{\rho}{4 M} \sum_{i}^M \sum_{j}^M ||x_i - x_j||_2^2
    \end{align}
    Finally we consider the following equivalent function in our optimization problem introducing $A = (a_{ij})_{i,j}$ a random matrix:
    \begin{align}
        \label{lossfunction3}
        \sum_{i=1}^M & \E\bigg[\lo(x_i, y) + \sum_{j}^M a_{ij}||x_i - x_j||_2^2 \bigg]  
    \end{align}
    In our gossip method the terms in the outer sum of (\ref{lossfunction3}) are sampled concurrently by different workers. $a_{ij}$ is a random variable controlling exchanges between workers $i$ and $j$ with $p = \mathds{P}(a_{ij} \neq 0)$ and $\E(a_{ij}) = \frac{\rho}{4 M}$.
        
    \subsection{GoSGD algorithm}
        The GoSGD algorithm considers $M$ independent agents called workers. Each of them hosts a CNN of the same architecture with a sets of weights noted $x_i$ for worker $i$. They are all initialized with the same value. During training all workers iteratively proceed two steps described below. One consisting on local optimisation with gradient descent and the other aiming at exchanging information in order to ensure a consensus between workers:
        
        \paragraph{Step 1 (Gradient update):}At all iterations $t$ a worker updates its hosted network's weights with a stochastic gradient descent on a random mini-batch. For the i-th worker the update is:
            $$ x_i^{t^+} = x_i^{t} - \eta^t v_i^t $$
        Where $\eta^t$ is the learning rate at iteration $t$ and $v_i^t= \frac{1}{|b(i,t)|}\sum_{y \in b(i,t)}\triangledown_x \lo(x_i^t, y)$ is an approximation computed on the sampled mini-batch $b(i,t)$ of the gradient of the expected error function at point $x_i^t$.
        \paragraph{Step 2 (Mixing update):}After the gradient descent each worker draws a random Bernoulli variable noted $S$ with expectancy $p$. This variable decides if the worker is sharing its information with another worker which will be chosen uniformly among the others.  To share the information between the update processes, we use a sum-weight gossip protocol~\cite{kempe2003gossip}. Sum-weight protocols use a sharing variable associated with each worker (noted $\alpha_i$ for agent $i$ and initialized to $\frac{1}{M}$) that is updated whenever information is exchanged and defines the rate at which information is mixed. Due to their push only nature, no synchronization is required between workers. The exchange between a worker $i$ drawing a successful $S$ and worker $j$ are described in algorithm 2.

        \begin{figure}[ht]
          \begin{minipage}[t]{.45\textwidth}
            \vspace*{-\baselineskip}
            \begin{algorithm}[H]
              \caption{GoSGD: workers Pseudo-code}\label{alg:a}
              \begin{algorithmic}[1]
                \State \textbf{Input:} $p$: probability of exchange, $M$: number of threads, $\eta:$ learning rate
                \State \textbf{Initialize:} $x$ is initialized randomly, $x_i = x$, $\alpha_i = \frac{1}{M}$ 
                \Repeat 
                    \State {\sc processMessages}(msg$_i$)
                	\State $x_i \leftarrow x_i - \eta^t v_i^t$     
                    \If{$S\sim B(p)$}
                        \State $j = Random(M)$
                        \State {\sc pushMessage}(msg$_j$)
                    \EndIf            
                \Until{Maximum iteration reached}    
                \State \Return $\frac{1}{M} \sum_{m=1}^M x_m $   
              \end{algorithmic}
            \end{algorithm}
          \end{minipage}%
          \hfill\vrule\hfill
              \begin{minipage}[t]{.45\textwidth}
                \vspace*{-\baselineskip}
                \begin{algorithm}[H]
                  \caption{Gossip update functions}\label{alg:b}
                  \begin{algorithmic}[1]
                    \Function{pushMessage}{queue msg$_j$}
                    \State $x_i \leftarrow x_i$
                    \State $\alpha_i \leftarrow \frac{\alpha_i }{2}$
                    \State msg$_j$.push(($x_i, \alpha_i$))
                    \EndFunction
                    \Function{processMessages}{queue msg$_i$}
                    \Repeat
                        \State $(x_j, \alpha_j) \leftarrow$ msg$_i$.pop()
                        \State $x_i \leftarrow \frac{\alpha_j}{\alpha_i + \alpha_j}x_j +  \frac{\alpha_i}{\alpha_i + \alpha_j}x_i$
                        \State $\alpha_i \leftarrow \alpha_j + \alpha_i$
                    \Until{msg$_i$.empty()}
                    \EndFunction
                  \end{algorithmic}
                \end{algorithm}
              \end{minipage}
        \end{figure}

        At each iteration a worker sends at most once its weights but can receive weights from several others. In this case the worker updates its weights sequentially in the reception order before performing any gradient update. Since agent $i$ can perform the update without waiting for an answer from $j$, and $j$ performs its update in a delayed fashion, no agent is ever idling and all computing resources are always being used (either performing a gradient update or a mixing update).
        
        Remark that these update rules are equivalent to common sum-weight gossip rules, with the main difference being that we choose not to scale $x_i$ which results in a more complex update rule for $x_j$. Consequently, several key properties of sum-weight protocols are retained:
        \paragraph{Property 1:}$\pmb{\alpha}^t = (\alpha_i^t)_{i=1..M}$ stays a stochastic vector.
        \paragraph{Property 2:}Consensus ($\forall i, x_i^t \rightarrow 1/M\sum_jx_j^t$) is obtained at exponential speed with respect to the number of mixing updates, when there is no gradient update.
        
        \paragraph{Remark:}$p$ is the only adjustable parameter of the algorithm. Obviously the bigger is $p$ the more exchanges between threads there are and eventually the closer the workers' weights will be. In our experiment, a low $p$ such as 0.01 already ensures a very good consensus.
        
    \subsection{Test model}
        The model that is evaluated on the test set is called test model. In the GoSGD method it is simply the averaging of all workers models weights:
            $$\overline{x}^t = \frac{1}{M}\sum_{i=1}^M x^{t}_i$$      
        
        It is possible to show that the test model after iteration $t$ can be rewritten:
            $$ \overline{x}^{t+1} = \overline{x}^t - \eta^t \sum_{i=1}^{M}\lambda_i^t v_i^t$$
        where $\pmb{\lambda}^t = (\lambda_i^t)_{i=1..M}$ is a stochastic vector with no null
        value.\\
        
        \paragraph{Need for a consensus:}In order that $v_i^t = \frac{1}{|b(i,t)|}\sum_{y \in b(i,t)}\triangledown_x \lo(x_i^t, y)$ contributes to the optimization of $\overline{x}^t$ it must be close to the direction of the true gradient at point $\overline{x}^t$. The best way to ensure this property is to keep $x_i^t$ and $\overline{x}^t$ as close as possible. This requirement holds for all workers at all iteration $t$. The initial optimization problem is thus coupled with a consensus problem.        
        \paragraph{Better gradient approximation:}If all workers all close enough to the consensus the averaging $\sum_{i=1}^{M}\lambda_i^t v_i^t$ result in a better approximation of the stochastic gradient as it is a weighted Monte Carlo indicator using $M$ times the number of draw than a single threaded optimization. 
                    
\section{Experiments} 
    We compare the convergence speed of GoSGD (gossip) with EASGD (elastic). The version of EASGD is the version with momentum (=0.99) namely MEASGD with parameters $\alpha = 0.887$ as suggested by the author. The parameter $\tau$ is equivalent to the inverse of our parameters $p$ and controls the frequency of exchange for one worker. The experiments have been done on CIFAR-10, see \cite{cifar} for detailed presentation. The network is the same used in \cite{elastic} and described in \cite{baseline} with a log loss. For the data sampling (loading and augmentation) we use the same protocol as in \cite{elastic}. During the training the learning rate is constant equal to 0.01 and the weight decay to $10^{-4}$. All batch contain 128 images. We use eight workers. The only adjustable parameter is the probability $p$ that control the frequency of exchange. We implemented both methods in torch framework \cite{torch} and we use 4 Titan x GPU.  
    
\begin{figure}
        \begin{tabular}{cc}
    			\includegraphics[width=0.5\textwidth]{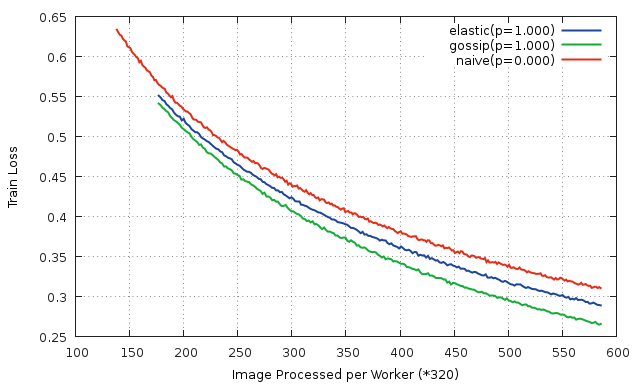}  &  \includegraphics[width=0.5\textwidth]{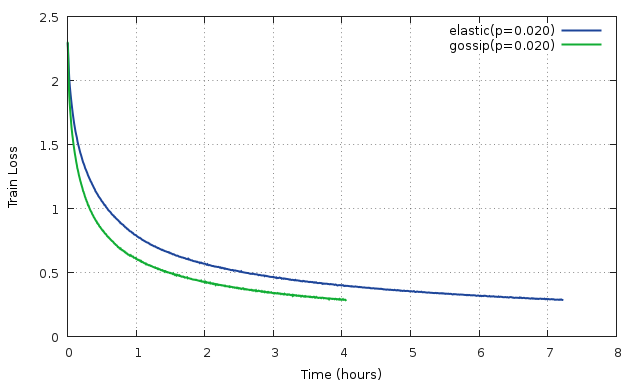}\\
    	\end{tabular}
       \caption{Evolution of the loss during training}
       \label{fig:losses}
    \end{figure}

    We report on Figure \ref{fig:losses} the evolution of the training losses for the different methods. As baseline we displayed a "Naive" scheme corresponding of a train without any exchange between workers. The train loss depicted is an averaging of the train loss of the last 50 batches taken regardless of the workers.
    
    The first curve on the left of Figure \ref{fig:losses} shows the loss against the number of images processed by each worker. For clarity, we zoomed the end of the convergence. To study the benefits of exchanging information regardless of the communication time, we maximize the number of exchanges by setting $p$ to 1. We can see that GoSGD do a better use of the exchanges than EASGD. It can signify that the gossip strategy implies a better consensus during training. 

    The second graph represents the evolution of the loss against time in hours. We use a small $p$ (0.02) as it seems to give a good compromise between communication costs and consensus both for GoSGD and EASGD. We can see that GoSGD is a lot faster than EASGD. Our strategy is converging in about 4 hours when EASGD needs more than 7 hours to reach the same train loss score. This shows that distributing SGD can benefit a lot from asynchronous strategies.

\section{Discussion} 
    In this paper, we have introduced a new learning scheme for deep architectures based on Gossip: GoSGD. We have experimentally validated our approach. Our algorithm disposes of several advantages compared to other methods. First, it is fully asynchronous and decentralized avoiding all kind of idling, then the exchanges are pairwise and benefit of the faster communication channel CPI. Second, there are theoretical aspects interesting to discuss: it is possible to derive a consensus convergence rate for many gossip algorithms. It could be useful to extend this study to GoSDG in order to measure the sensibility of gossip averaging to the additional gradients. This would provide some insights to optimize the frequency of exchange and to control it as low as possible without impacting too much the consensus between threads.

\medskip

\bibliographystyle{acm}
\bibliography{biblio}
\end{document}